\documentclass[nohyperref]{article}

\usepackage{microtype}
\usepackage{graphicx}
\usepackage{booktabs} 

\usepackage{hyperref}

\usepackage[accepted]{icml2022}

\usepackage{amsmath}
\usepackage{amssymb}
\usepackage{mathtools}
\usepackage{amsthm}

\usepackage[capitalize,noabbrev]{cleveref}

\theoremstyle{plain}

\theoremstyle{definition}

\theoremstyle{remark}

\usepackage[textsize=tiny]{todonotes}

\icmltitlerunning{Zero-Shot Sketch-to-Image Synthesis}

\begin{document}

\twocolumn[
\icmltitle{Style-Content Disentanglement in Language-Image Pretraining Representations for Zero-Shot Sketch-to-Image Synthesis}



\icmlsetsymbol{equal}{}

\begin{icmlauthorlist}
\icmlauthor{Jan Zuiderveld}{yyy}
\end{icmlauthorlist}

\icmlaffiliation{yyy}{University of Amsterdam}

\icmlcorrespondingauthor{Jan Zuiderveld}{janzuiderveld@gmail.com}

\icmlkeywords{Style-Content Disentanglement, Sketch-to-Image Synthesis, Language-Image Pretraining}

\vskip 0.3in
]



\printAffiliationsAndNotice{}  

\begin{abstract}
In this work, we propose and validate a framework to leverage language-image pretraining representations for training-free zero-shot sketch-to-image synthesis. We show that disentangled content and style representations can be utilized to guide image generators to employ them as sketch-to-image generators without (re-)training any parameters. Our approach for disentangling style and content entails a simple method consisting of elementary arithmetic assuming compositionality of information in representations of input sketches. Our results demonstrate that this approach is competitive with state-of-the-art instance-level open-domain sketch-to-image models, while only depending on pretrained off-the-shelf models and a fraction of the data.


\end{abstract}

\begin{figure*}
  \includegraphics[width=\textwidth]{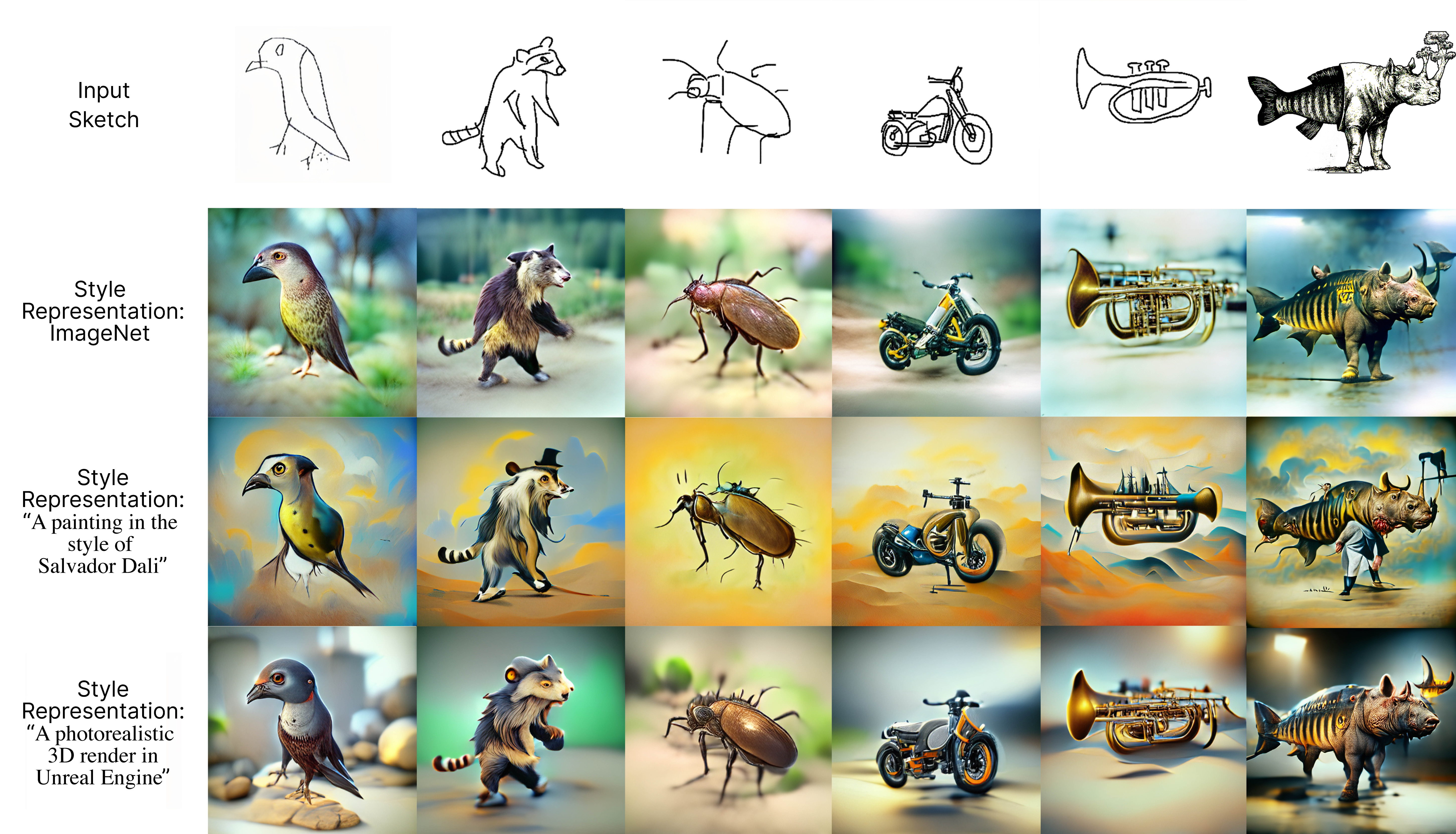}
  \caption{Several selected zero-shot outputs of our proposed framework, with given input sketches and style representations.}
  
  \label{fig:teaser}
\end{figure*}

\section{Introduction}






The power and promise of deep generative models such as GANs \cite{goodfellow2014generative} lie in their ability to synthesize endless realistic, diverse, and novel content with minimal user effort. The potential utility of these models continues to grow thanks to the increased quality and resolution of large-scale generative models in recent years. 
Conditional image synthesis allows users to use inputs to control the output of image synthesis methods. A variety of input modalities have been studied, mostly based on conditional GANs \cite{mirza2014conditional}. Several methods exist that allow generative models to be guided by conditioning variables, e.g. text, sketches or segmentation maps. Sketch conditioning finds a practical middle ground within the characteristics of these methods. It allows significantly more control over output structure than text conditioning, while not requiring labels as segmentation conditioning does. \\


However, current sketch conditioned models are trained on a relatively narrow set of visual concepts. This restricts the space of possible inputs that result in desired output images, impeding the general usability of these models. An important cause of this is the fact that paired image datasets are labor intensive and costly to create. Approaches that circumvent this requirement exist, including CycleGAN  based architectures \cite{liuUnsupervisedSketchtoPhotoSynthesis2020, xiangAdversarialOpenDomain2021} and the usage of image-to-sketch models for synthesizing paired data \citestyle{liuSelfSupervisedSketchtoImageSynthesis2020, huangMultimodalConditionalImage2021}. But as of yet, no work has been published applying sketch-to-image synthesis on large, highly diverse datasets.

Recently, the idea of learning visual concepts from supervision contained in natural language has gained a lot of attention. Language-Image Pre-training models learn unified visual- and natural language representations from the supervision contained in the vast amount of text and associated images on the internet, resulting in very rich visual representations and strong zero-shot classification performance. 





In this work, we adopt a framework to leverage classifier representations for sketch-to-image generation. A simple method consisting of elementary arithmetic assuming compositionality of information in the representations is used to disentangle spatial and conceptual information from stylistic information in representations of input sketches. We show that these disentangled representations can then be utilized to guide image generators to employ them as sketch-to-image generators without (re-)training any parameters.\footnote{Our code and additional samples will be made publicly available at [retracted for anonymity]} 

Our results demonstrate that this approach, when used with sufficiently flexible image generators and informative representations, is competitive with state-of-the-art instance-level open-domain sketch-to-image models evaluated on in-domain inputs. Current state-of-the-art models feature complex architectures and require large datasets, while our proposed framework depends on pretrained off-the-shelf models, requires no training, and is evaluated in a zero-shot fashion. Figure \ref{fig:teaser} shows several selected sketch-to-image translations by our approach. 

\bigskip

Our approach is simple to implement while also requiring minimal tuning. To emphasize the potential of the core methodology itself, we report results with default hyperparameters and no additional optimizations. The effectiveness of our method indicates that representations learned by language-image pretraining exhibit strong compositionality, something that has not been harnessed for disentanglement in earlier research.

\section{Related Work}
\subsection{Sketch to Image Synthesis}
The goal of sketch-based image synthesis is to output a target image from a given sketch. Early works \cite{chen2009sketch2photo, chen2012poseshop, eitz2011photosketcher} regard freehand sketches as queries or constraints to retrieve each composition and stitch them into a picture, requiring availability of close image matches in the queried database for reasonable results. In recent years, an increasing number of works adopt GAN-based models to learn sketch-to-image synthesis directly. 

Several works \cite{zhu2017unpaired, li2019linestofacephoto, chen2020deepfacedrawing} train Conditional GANs  with photos and corresponding edge maps to make up for the lack of real sketch data, but only report results trained on single class datasets.  However, sketches differ from edge maps in several ways, resulting in inferior results when applying these models to sketches. The authors of SketchyGAN \cite{chenSketchyGANDiverseRealistic2018} extended the largest currently available paired dataset of sketches and pictures (75k sketches across 125 categories), extended it and trained a conditional GAN on these. ContextualGAN \cite{luImageGenerationSketch2018} turns the image generation problem into an image completion problem: the network learns the joint distribution of sketch and image pairs and acquires the result by iteratively traversing the manifold. PoE-GAN \cite{huangMultimodalConditionalImage2021} is a multimodal conditional GAN that is trained on images with paired text, sketch and segmentation maps, achieving state-of-the-art scene-level sketch-to-image synthesis. Xiang et al. \cite{xiangAdversarialOpenDomain2021} propose a framework that jointly learns sketch-to-photo and photo-to-sketch generation to help in generalizing to open-domain classes, and achieve state-of-the-art instance-level sketch-to-image synthesis.

All of these previously proposed architectures are trained on datasets with a limited amount of classes. To the best of our knowledge, there is no published research reporting training any kind of image-to-image translation architectures on large-scale, diverse datasets for zero-shot image-to-image, or sketch-to-image synthesis.



\subsection{Language-Image Pre-training}
Multiple recent works learn cross-modal vision and language representations \cite{lu2019vilbert, xu2018attngan} for a variety of tasks. Following the success of  Transformers \cite{vaswaniAttentionAllYou2017} in various language tasks, recent vision and joint vision language methods typically use transformers as their backbone. A recent model, based on Contrastive Language-Image Pre-training (CLIP) \cite{radfordLearningTransferableVisual2021}, consisting of a transformer- language encoder and image encoder is trained to learn a multi-modal representation space. which can be used to estimate the semantic similarity between a given text and an image by evaluating embeddings' cosine similarity.

CLIP was trained on 400 million text-image pairs, collected from a variety of publicly available sources on the Internet. The representations learned by CLIP have been shown to be extremely powerful, enabling state-of-the-art zero-shot image classification on a variety of datasets.

\begin{figure*}[!h] 
  \includegraphics[width=\textwidth]{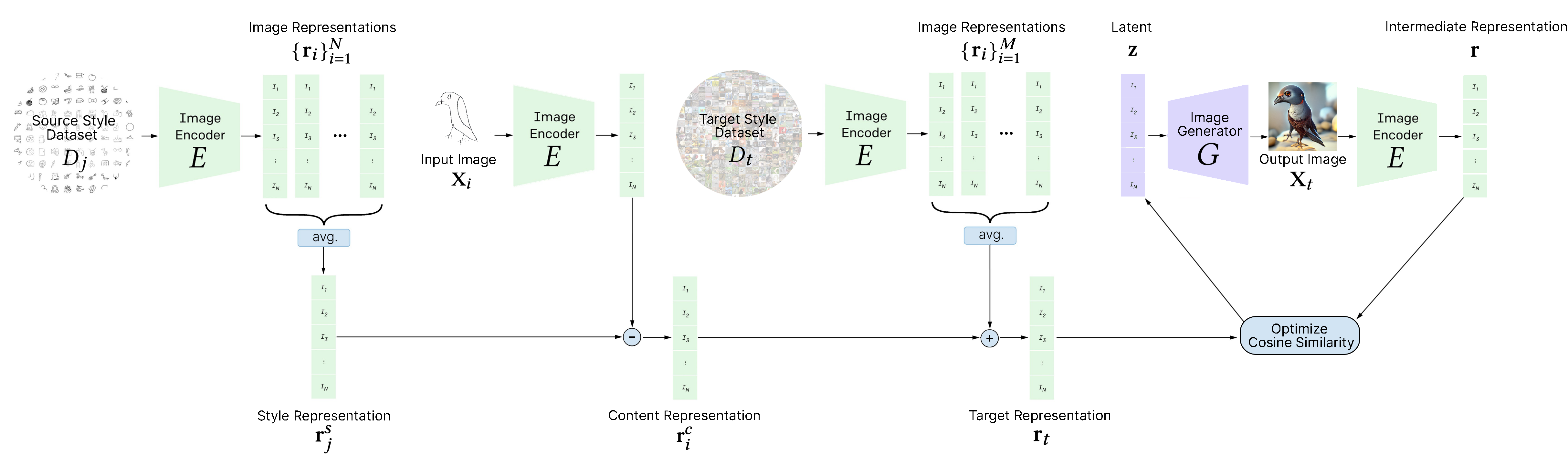}
  \caption{Overview of our proposed framework.}
  
  \label{fig:overview}
\end{figure*}

\subsection{Classifier Guidance Based Image Synthesis}
GANs for conditional image synthesis make heavy use of class labels. This often takes the form of class-conditional normalization statistics \cite{de2017modulating} as well as discriminator heads that are explicitly designed to behave like classifiers \cite{miyato2018cgans}, indicating that class information is crucial to the success of these models. 
The general idea of optimizing  latent representations or parameters of an image generator using a separately trained classifier has been widely used as a powerful framework for generating, editing and recovering images \cite{ dhariwal2021diffusion,  abdal2019image2stylegan, harkonen2020ganspace}. Santurkar et al. \cite{santurkar2019image} show that adversarial robustness of the classifier is a crucial aspect for optimal execution of such tasks.

Several projects use CLIP as a classifier to guide text-to-image generation through optimization. Examples include optimizing the weights of implicit neural representation networks \cite{stanleyCompositionalPatternProducing2007}, the latent space of StyleGAN2 \cite{karras2020analyzing} and VQGAN \cite{esser2021taming}, and guiding diffusion architectures \cite{nichol2021glide}.

\section{Methodology}

Much current research in unsupervised image-to-image translation relies on learning content and style representations. In these approaches, the content $\mathbf{c}_i$ of input images $\mathbf{X}_i$ and the style of target images $\mathbf{s}_K$ together condition a generator G that synthesizes translated images $\mathbf{X}_t = G( \mathbf{c}_i, \mathbf{s}_t)$. 

Instead of learning content and style representations from scratch, we propose a simple method for disentangling style and content in representations of a pretrained image encoder $E$ under the assumptions of in-distribution input and compositionality of representations. We use obtained representations to guide $G$.  We apply this framework to zero-shot sketch-to-image synthesis. To emphasize the potential of the core methodology itself, we report results with default hyperparameters and no additional optimizations. See Figure \ref{fig:overview} for a general overview. 

\subsection{Problem Formulation}
Let us model the formation of an image $\mathbf{X}$ as a function of style $\mathbf{s} \in \mathcal{S}$ and content $\mathbf{c} \in \mathcal{C}$, $\mathbf{X} = G\left(\mathbf{s}, \mathbf{c}\right)$.  Let $\mathcal{P}$ denote the set of possible "parts" of an image, these parts be any feature of the input, e.g. the presence and location of an object, features of this object such as its size and color, or global style of an image. Let $\mathcal{X}$ denote the input space. For each $\mathbf{X} \in \mathcal{X}$ , we assume the existence of a function $D$ mapping $\mathbf{X} \text { to } \mathcal{P}^{\prime} \subseteq \mathcal{P}$, the set of its parts. Style $\mathbf{s}$ describes all parts $\mathbf{p}^s \in D(\mathbf{X})$ that are invariant within the stylistic domain of $\mathbf{X}$, while content $\mathbf{c}$ describes all residual parts $\mathbf{p}^c \in D(\mathbf{X})$ that are not described by style $\mathbf{s}$.  

Given an input dataset $D_j$ of $N$ images $\mathbf{X}_i$, all sampled from stylistic domain $\mathbf{s}_j$ with normally distributed content $\mathbf{c}_i$,
$$
D_j=\left\{X_{i} \right\}_{i=1}^{N}, \quad \mathbf{X}_i = G\left(\mathbf{s}_j, \mathbf{c}_i\right), \quad \mathbf{c}_i \sim \mathcal{N}\left(0, \sigma^{2} \mathbf{I}\right),
$$
\noindent we use image encoder $E$: $\mathcal{X} \rightarrow \mathcal{R}$ to  obtain representations $\mathbf{r}_i = E(\mathbf{X}_i)$, where $\mathcal{R}$ denotes the representation space. We assume $\mathbf{r}_i$ to be compositional, meaning it can be expressed as a weighted sum of simpler parts. Let $h: \mathcal{P} \rightarrow \mathcal{R}$ denote a function that maps parts to representations. Formally, we define a function $f(\mathbf{X}) \in \mathcal{R}$ as compositional if it can be expressed as a weighted sum of the elements of $\{h(p) \mid p \in D(\mathbf{X})\}$

Then, we can decompose $\mathbf{r}_i$ as a sum of content representation $\mathbf{r}_i^c$ and style representation $\mathbf{r}_j^s$, $\mathbf{r}_i = \mathbf{r}_i^c +  \mathbf{r}_j^s$. We assume $\mathbf{X}_i$ to be in-distribution for $E(\mathbf{X})$, and note that $\mathbf{r}_i^c$ follows the distribution of $\mathbf{c}_i$. We can then obtain $ \mathbf{r}_j^s$ by taking the arithmetic mean over all $\mathbf{r}_i$, 
$$ \mathbf{r}_j^s =\frac{1}{N} \sum_{i=1}^{N} \mathbf{r}_i = \frac{1}{N} \sum_{i=1}^{N} r^\mathbf{c}_i + \frac{1}{N} \sum_{i=1}^{N} \mathbf{r}_j^s =  0 + \frac{N}{N} \mathbf{r}_j^s.$$ 
\noindent Finally, to obtain $\mathbf{r}_i^c$ we subtract $\mathbf{r}_j^s$ from $\mathbf{r}_i$,
$$\mathbf{r}_i^c = \mathbf{r}_i - \mathbf{r}_j^s = \mathbf{r}_i^c + \mathbf{r}_j^s  - \mathbf{r}_j^s.$$ 
\noindent Using these operations we obtain $\mathbf{r}_i^c$ describing the content of input images $\mathbf{X}_i$ from stylistic domain $\mathbf{s}_j$, and $ \mathbf{r}_{t}^s$ describing target stylistic domain $\mathbf{s}_t$ of dataset $D_t$ of $M$ images. We sum these to create target representations for guiding image synthesis: $\mathbf{r}_t = \mathbf{r}_i^c + \mathbf{r}_t^s $.

Specifically, given image generator $G(\mathbf{z})$,  and image encoder $E$, we solve the following optimization problem to iteratively update $\mathbf{X}_t$, a translation of $\mathbf{X}_i$ in style $\mathbf{s}_t$:
$$\underset{\mathbf{z} \in \mathcal{Z}}{\arg \min } \frac{\left\langle E(G(\mathbf{z})), \mathbf{r}_t \right\rangle}{\left\|E(G(\mathbf{z})) \right\| \cdot\left\|\mathbf{r}_t\right\|},$$
\noindent where $\langle\cdot, \cdot\rangle$ computes the cosine similarity between its arguments.

\begin{figure*}
  \centering
  \includegraphics[width=0.73\textwidth]{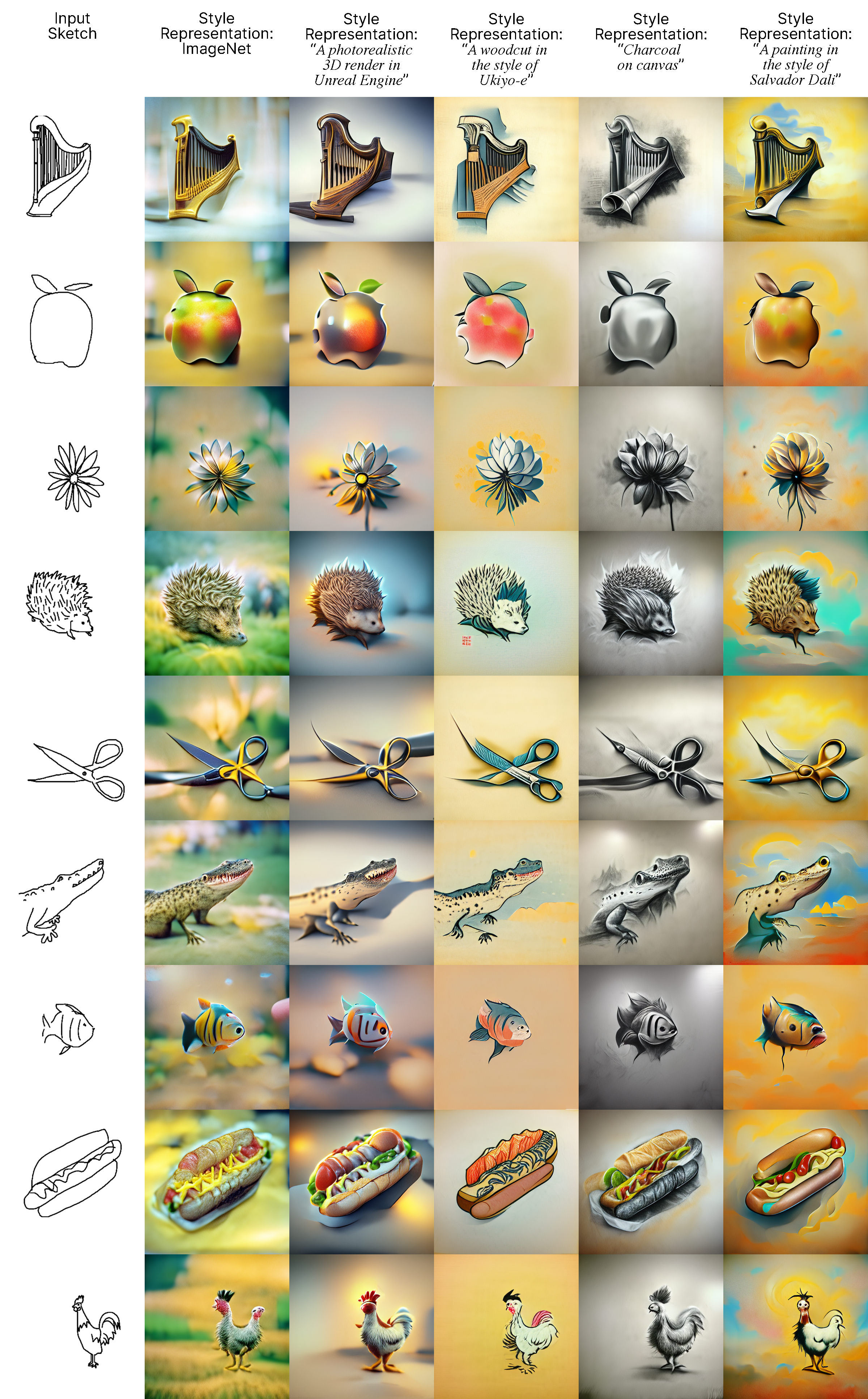}
  \caption{Zero-shot sketch-to-image results $\mathbf{X}_t$ for several random source images $\mathbf{X}_i$ and target styles $\mathbf{s}_t$}. 
  
  \label{fig:large_grid}
\end{figure*}

\subsection{Improving Adversarial Robustness}
Most off-the-shelf classifiers are not trained for exhibiting adversarial robustness. Since we are directly optimizing images using such classifiers, this process is prone to induce non-semantic perturbations in $\mathbf{X}_t$ that increase classification scores \cite{liuFuseDreamTrainingFreeTexttoImage2021}. Inspired by self-supervised learning methods, we use augmentation pipeline $a$ when feeding input $\mathbf{X}_t$ to image encoder $E$:
$$
a(\mathbf{X}_t) =\mathbb{E}_{\mathbf{X}_t^{\prime} \sim \pi(\cdot \mid \mathbf{X}_t)}
$$
where $\mathbf{X}_t^{\prime}$ is a random perturbation of the input image $\mathbf{X}_t$ drawn from distribution $\pi(\cdot \mid \mathbf{X}_t)$ of candidate data augmentations, including random colorization, translation, and cutout. This generates new samples $\mathbf{X}_t^{\prime}$ that should average out adversarial gradients while preserving content and style information in $\mathbf{X}_t$.

\section{Experimental setup}
\subsection{Image Encoder $\mathbf{E}$}
To evaluate the adopted framework we report experiments with CLIP \cite{radfordLearningTransferableVisual2021} as the image encoder $E$. CLIP is well-suited for our framework as the large-scale dataset it has been trained on ensures that a wide range of styles $\mathbf{s}$ are in-distribution. And, more importantly, the excellent zero-shot performance of CLIP indicates its representations exhibit a significant amount of compositionality, making it well-suited for our style-content disentanglement method. 

The notion that zero-shot classification performance strongly correlates with compositionality in learned representations is demonstrated by Sylvain et al. \cite{sylvain2019locality}. This is very intuitive: we expect compositionality to be an advantage in zero-shot learning: if a model has a good understanding of how parts map to representations, it can learn to combine known concepts to describe new classes.

Since CLIP also includes a natural language encoder which maps text to the same embedding space as its image encoder, we also experiment with target styles $\mathbf{s}_t$ obtained through directly encoding text.

\subsection{Image Generator $\mathbf{G}$}
The space of images our approach is able to synthesize is limited by the representational power of the image generator we use. Our method is prone to inherit biases of the used generator, either induced by its training data e.g. center, spatial or color bias, or by inductive biases present in its architecture.

Since we aim to achieve zero-shot sketch-to-image synthesis, the image generator needs to be relatively flexible. However, very flexible image generators such as differentiable image parameterizations, e.g. implicit neural representations \cite{stanleyCompositionalPatternProducing2007} or direct pixel optimization, facilitate adversarial attacks. 

Based on preliminary experiments we use VQGAN \cite{esser2021taming} as $G$ for our main experiments. VQGAN exhibits flexible image synthesis, arguably due to its latent embedding structure. We initialize $\mathbf{z}$ by using VQGAN's image encoder to encode input sketch $\mathbf{X}_i$. 

In \ref{sec:ablations_glide} we report results of using GLIDE \cite{dhariwal2021diffusion} as $G$, a recently proposed diffusion model trained on a similar dataset as CLIP. We use a small version of this model, trained with filtered data, since the weights of the full model have not been released. In these experiments the model is guided by representations obtained by a smaller CLIP model trained on this same filtered dataset, with noise augmentations.

\subsection{Datasets}
For obtaining sketch style representations $\mathbf{s}_j$, we use sketches of the sketchy dataset \cite{sangkloy2016sketchy}, a collection of 75,471 sketch-photo pairs sampled from 125 categories. When evaluating our framework we sample 10k images from the sketchy dataset to determine $\mathbf{s}_j$, excluding the class of the input sketch. Thus, all reported image translations can be considered zero-shot outputs. In section \ref{sec:ablations_data} we report results of using 2, 10, 100, 1k or 70k samples from the sketchy dataset for determining $\mathbf{s}_j$. For obtaining target style $\mathbf{s}_t$ we use 10k randomly sampled items of ImageNet \cite{deng2009imagenet}, unless specified otherwise. 



\section{Results}




%


\label{sec:comp}

Figure \ref{fig:large_grid} shows sketch-to-image results $\mathbf{X}_t$ for several random source images $\mathbf{X}_i$ and target styles $\mathbf{s}_t$ obtained from ImageNet and several text prompts.

\begin{figure}
  \includegraphics[width=0.5\textwidth]{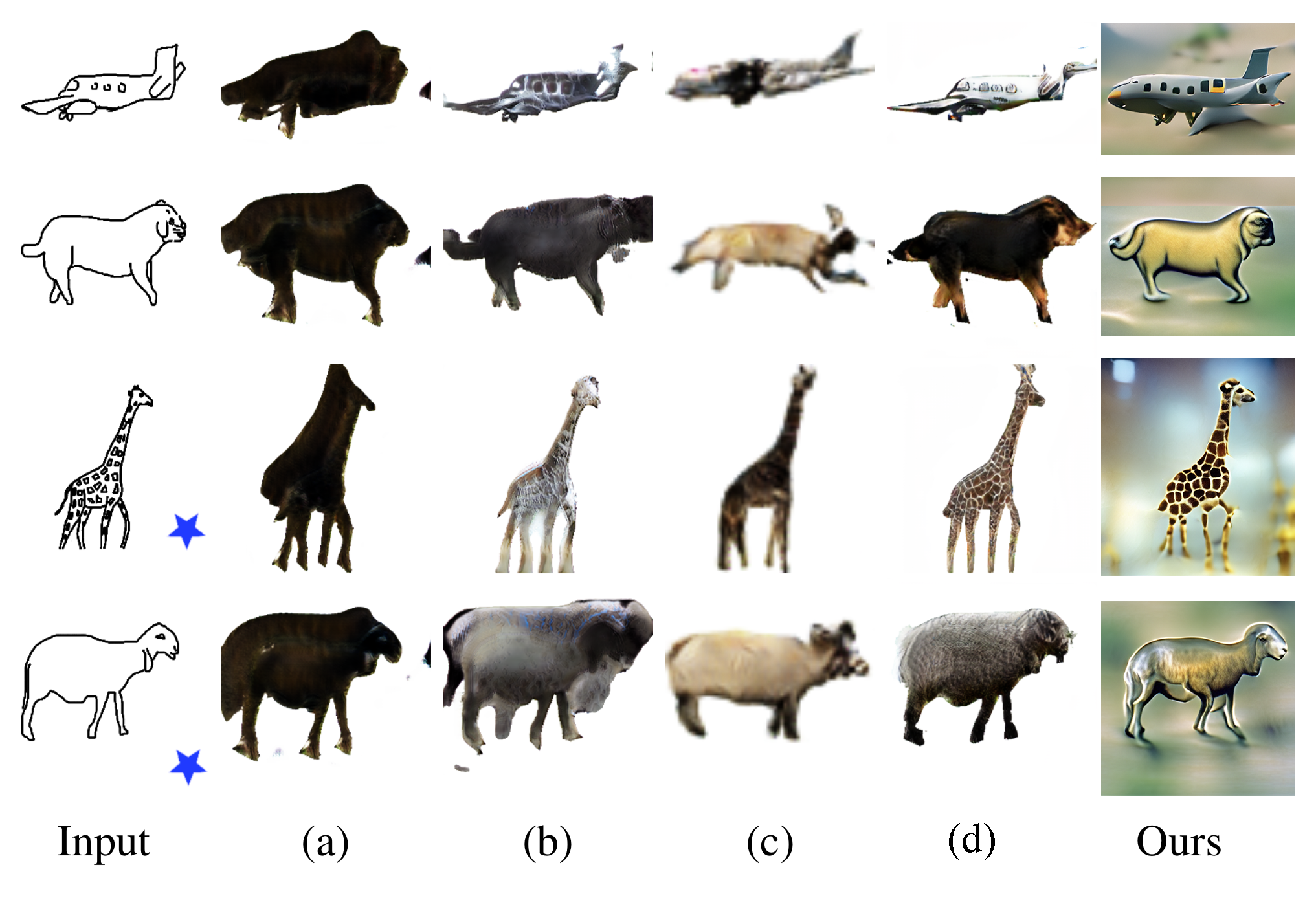}
  \caption{Sketch-to-image results of several methods, (a) CycleGAN \cite{zhuUnpairedImagetoImageTranslation2017}, (b) conditional
CycleGAN, (c) EdgeGAN \cite{gaoSketchyCOCOImageGeneration2020}, (d) Xiang et al. \cite{xiangAdversarialOpenDomain2021} and ours. Class conditioned inputs are marked \textit{without} a star. Note that our framework does not make use of class information, and is the only one that is evaluated in a zero-shot fashion.}
  
  \label{fig:comp}
\end{figure}

To better illustrate the effectiveness of our proposed solution, we show some quick and dirty comparisons to state-of-the-art open domain sketch-to-image synthesis by Xiang et al. \cite{xiangAdversarialOpenDomain2021}. In their work, Xiang et al. report qualitative comparisons of their results with CycleGAN \cite{zhuUnpairedImagetoImageTranslation2017}, conditional CycleGAN, and EdgeGAN \cite{gaoSketchyCOCOImageGeneration2020}. We fed all compared input sketches in their work to our proposed framework and added the outputs to their comparison, see Figure \ref{fig:comp}. Entries not marked with a star are conditioned on their respective class. Our framework shows competitive results without making use of class information, and while being the only method evaluated in a zero-shot fashion.

\subsection{Ablations}
\label{sec:ablations}

\subsubsection{Varying Dataset Size for Obtaining $\mathbf{s}_{j}$}
\label{sec:ablations_data}
\begin{figure}[!h]
  \includegraphics[width=0.5\textwidth]{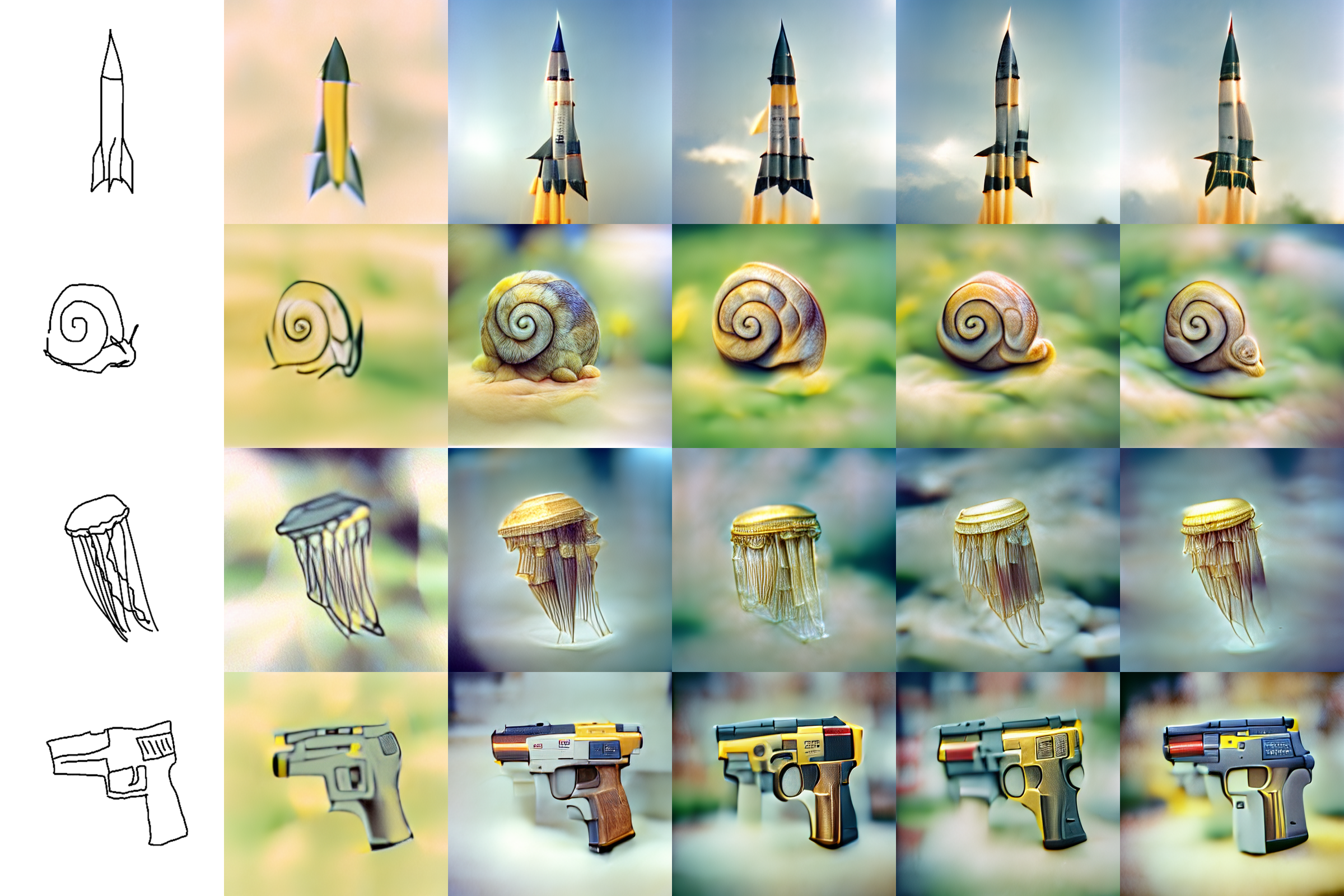}
  \caption{Zero-shot sketch-to-image results $\mathbf{X}_t$ for several source images $\mathbf{X}_i$ with a varying amount of samples from the sketchy dataset for determining $\mathbf{s_j}$ and 10k samples from ImageNet for obtaining target style $\mathbf{s}_t$. From left to right: input sketch, 2, 10, 100, 1k and 70k.}
  
  \label{fig:dataset}
\end{figure}

To shed light on the required dataset size we report results of varying the amount of used samples (2, 10, 100, 1k, 70k) for calculating $\mathbf{s}_j$, see Figure \ref{fig:dataset}. Remarkably, 10 samples are enough to get decent results, and after using more than 100 samples gains are minimal. 

\subsubsection{Using GLIDE as ${G}$} 
\label{sec:ablations_glide}
\begin{figure}[!h]
  \includegraphics[width=0.5\textwidth]{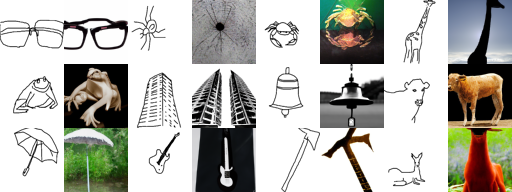}
  \caption{Sketch-to-image results $\mathbf{X}_t$ using GLIDE as $G$ for several source images $\mathbf{X}_i$ }
  
  \label{fig:GLIDE}
\end{figure}

In Figure \ref{fig:GLIDE} we show a handful of random results of employing GLIDE \cite{dhariwal2021diffusion} as $G$ instead of VQGAN \cite{esser2021taming}. We tried guiding GLIDE with different target styles $\mathbf{s}_t$, but this was not effective. A possible reason for this could be the fact that representations learned by the smaller, noised CLIP model used in GLIDE show less compositionality and/or contain less information than the regular CLIP model released by OpenAI.  

\section{Conclusion}

We show that simple arithmetic can be an effective tool for intuitively \textit{editing} language-image pretraining architectures' learned representations, and prove this to be a computationally cheap method for disentangling style and content. 

Our method applies this idea to achieve zero-shot instance-level sketch-to-image synthesis competitive with state-of-the-art instance-level open-domain sketch-to-image models, while only depending on pretrained off-the-shelf architectures and a sketch dataset as small as 100 items. 

Additionally, these results suggest that large language-image pretraining architectures' learned representations show a significant amount of compositionality, similar to what is generally observed in word-embeddings \cite{gittensSkipGramZipfUniform2017, allenAnalogiesExplainedUnderstanding2019}. 

\bibliography{sample-base}

\begin{thebibliography}{35}
\providecommand{\natexlab}[1]{#1}
\providecommand{\url}[1]{\texttt{#1}}
\expandafter\ifx\csname urlstyle\endcsname\relax
  \providecommand{\doi}[1]{doi: #1}\else
  \providecommand{\doi}{doi: \begingroup \urlstyle{rm}\Url}\fi

\bibitem[Abdal et~al.(2019)Abdal, Qin, and Wonka]{abdal2019image2stylegan}
Abdal, R., Qin, Y., and Wonka, P.
\newblock Image2stylegan: How to embed images into the stylegan latent space?
\newblock In \emph{Proceedings of the IEEE/CVF International Conference on
  Computer Vision}, pp.\  4432--4441, 2019.

\bibitem[Allen \& Hospedales(2019)Allen and
  Hospedales]{allenAnalogiesExplainedUnderstanding2019}
Allen, C. and Hospedales, T.
\newblock Analogies {Explained}: {Towards} {Understanding} {Word} {Embeddings}.
\newblock \emph{arXiv:1901.09813 [cs, stat]}, May 2019.
\newblock URL \url{http://arxiv.org/abs/1901.09813}.
\newblock arXiv: 1901.09813.

\bibitem[Chen et~al.(2020)Chen, Su, Gao, Xia, and Fu]{chen2020deepfacedrawing}
Chen, S.-Y., Su, W., Gao, L., Xia, S., and Fu, H.
\newblock Deepfacedrawing: Deep generation of face images from sketches.
\newblock \emph{ACM Transactions on Graphics (TOG)}, 39\penalty0 (4):\penalty0
  72--1, 2020.

\bibitem[Chen et~al.(2009)Chen, Cheng, Tan, Shamir, and
  Hu]{chen2009sketch2photo}
Chen, T., Cheng, M.-M., Tan, P., Shamir, A., and Hu, S.-M.
\newblock Sketch2photo: Internet image montage.
\newblock \emph{ACM transactions on graphics (TOG)}, 28\penalty0 (5):\penalty0
  1--10, 2009.

\bibitem[Chen et~al.(2012)Chen, Tan, Ma, Cheng, Shamir, and
  Hu]{chen2012poseshop}
Chen, T., Tan, P., Ma, L.-Q., Cheng, M.-M., Shamir, A., and Hu, S.-M.
\newblock Poseshop: Human image database construction and personalized content
  synthesis.
\newblock \emph{IEEE Transactions on Visualization and Computer Graphics},
  19\penalty0 (5):\penalty0 824--837, 2012.

\bibitem[Chen \& Hays(2018)Chen and Hays]{chenSketchyGANDiverseRealistic2018}
Chen, W. and Hays, J.
\newblock {SketchyGAN}: {Towards} {Diverse} and {Realistic} {Sketch} to {Image}
  {Synthesis}.
\newblock In \emph{2018 {IEEE}/{CVF} {Conference} on {Computer} {Vision} and
  {Pattern} {Recognition}}, pp.\  9416--9425, Salt Lake City, UT, June 2018.
  IEEE.
\newblock ISBN 978-1-5386-6420-9.
\newblock \doi{10.1109/CVPR.2018.00981}.
\newblock URL \url{https://ieeexplore.ieee.org/document/8579079/}.

\bibitem[De~Vries et~al.(2017)De~Vries, Strub, Mary, Larochelle, Pietquin, and
  Courville]{de2017modulating}
De~Vries, H., Strub, F., Mary, J., Larochelle, H., Pietquin, O., and Courville,
  A.
\newblock Modulating early visual processing by language.
\newblock \emph{arXiv preprint arXiv:1707.00683}, 2017.

\bibitem[Deng et~al.(2009)Deng, Dong, Socher, Li, Li, and
  Fei-Fei]{deng2009imagenet}
Deng, J., Dong, W., Socher, R., Li, L.-J., Li, K., and Fei-Fei, L.
\newblock Imagenet: A large-scale hierarchical image database.
\newblock In \emph{2009 IEEE conference on computer vision and pattern
  recognition}, pp.\  248--255. Ieee, 2009.

\bibitem[Dhariwal \& Nichol(2021)Dhariwal and Nichol]{dhariwal2021diffusion}
Dhariwal, P. and Nichol, A.
\newblock Diffusion models beat gans on image synthesis.
\newblock \emph{arXiv preprint arXiv:2105.05233}, 2021.

\bibitem[Eitz et~al.(2011)Eitz, Richter, Hildebrand, Boubekeur, and
  Alexa]{eitz2011photosketcher}
Eitz, M., Richter, R., Hildebrand, K., Boubekeur, T., and Alexa, M.
\newblock Photosketcher: interactive sketch-based image synthesis.
\newblock \emph{IEEE Computer Graphics and Applications}, 31\penalty0
  (6):\penalty0 56--66, 2011.

\bibitem[Esser et~al.(2021)Esser, Rombach, and Ommer]{esser2021taming}
Esser, P., Rombach, R., and Ommer, B.
\newblock Taming transformers for high-resolution image synthesis.
\newblock In \emph{Proceedings of the IEEE/CVF Conference on Computer Vision
  and Pattern Recognition}, pp.\  12873--12883, 2021.

\bibitem[Gao et~al.(2020)Gao, Liu, Xu, Wang, Liu, and
  Zou]{gaoSketchyCOCOImageGeneration2020}
Gao, C., Liu, Q., Xu, Q., Wang, L., Liu, J., and Zou, C.
\newblock {SketchyCOCO}: {Image} {Generation} from {Freehand} {Scene}
  {Sketches}.
\newblock \emph{arXiv:2003.02683 [cs]}, April 2020.
\newblock URL \url{http://arxiv.org/abs/2003.02683}.
\newblock arXiv: 2003.02683.

\bibitem[Gittens et~al.(2017)Gittens, Achlioptas, and
  Mahoney]{gittensSkipGramZipfUniform2017}
Gittens, A., Achlioptas, D., and Mahoney, M.~W.
\newblock Skip {Gram}: {Zipf} plus {Uniform} equals {Vector} {Additivity}.
\newblock In \emph{Proceedings of the 55th {Annual} {Meeting} of the
  {Association} for {Computational} {Linguistics} ({Volume} 1: {Long}
  {Papers})}, pp.\  69--76, Vancouver, Canada, July 2017. Association for
  Computational Linguistics.
\newblock \doi{10.18653/v1/P17-1007}.
\newblock URL \url{https://aclanthology.org/P17-1007}.

\bibitem[Goodfellow et~al.(2014)Goodfellow, Pouget-Abadie, Mirza, Xu,
  Warde-Farley, Ozair, Courville, and Bengio]{goodfellow2014generative}
Goodfellow, I., Pouget-Abadie, J., Mirza, M., Xu, B., Warde-Farley, D., Ozair,
  S., Courville, A., and Bengio, Y.
\newblock Generative adversarial nets.
\newblock \emph{Advances in neural information processing systems}, 27, 2014.

\bibitem[H{\"a}rk{\"o}nen et~al.(2020)H{\"a}rk{\"o}nen, Hertzmann, Lehtinen,
  and Paris]{harkonen2020ganspace}
H{\"a}rk{\"o}nen, E., Hertzmann, A., Lehtinen, J., and Paris, S.
\newblock Ganspace: Discovering interpretable gan controls.
\newblock \emph{arXiv preprint arXiv:2004.02546}, 2020.

\bibitem[Huang et~al.(2021)Huang, Mallya, Wang, and
  Liu]{huangMultimodalConditionalImage2021}
Huang, X., Mallya, A., Wang, T.-C., and Liu, M.-Y.
\newblock Multimodal {Conditional} {Image} {Synthesis} with
  {Product}-of-{Experts} {GANs}.
\newblock \emph{arXiv:2112.05130 [cs]}, December 2021.
\newblock URL \url{http://arxiv.org/abs/2112.05130}.
\newblock arXiv: 2112.05130.

\bibitem[Karras et~al.(2020)Karras, Laine, Aittala, Hellsten, Lehtinen, and
  Aila]{karras2020analyzing}
Karras, T., Laine, S., Aittala, M., Hellsten, J., Lehtinen, J., and Aila, T.
\newblock Analyzing and improving the image quality of stylegan.
\newblock In \emph{Proceedings of the IEEE/CVF Conference on Computer Vision
  and Pattern Recognition}, pp.\  8110--8119, 2020.

\bibitem[Li et~al.(2019)Li, Chen, Wu, and Zha]{li2019linestofacephoto}
Li, Y., Chen, X., Wu, F., and Zha, Z.-J.
\newblock Linestofacephoto: Face photo generation from lines with conditional
  self-attention generative adversarial networks.
\newblock In \emph{Proceedings of the 27th ACM International Conference on
  Multimedia}, pp.\  2323--2331, 2019.

\bibitem[Liu et~al.(2020)Liu, Yu, and
  Yu]{liuUnsupervisedSketchtoPhotoSynthesis2020}
Liu, R., Yu, Q., and Yu, S.
\newblock Unsupervised {Sketch}-to-{Photo} {Synthesis}.
\newblock \emph{arXiv:1909.08313 [cs]}, March 2020.
\newblock URL \url{http://arxiv.org/abs/1909.08313}.
\newblock arXiv: 1909.08313.

\bibitem[Liu et~al.(2021)Liu, Gong, Wu, Zhang, Su, and
  Liu]{liuFuseDreamTrainingFreeTexttoImage2021}
Liu, X., Gong, C., Wu, L., Zhang, S., Su, H., and Liu, Q.
\newblock {FuseDream}: {Training}-{Free} {Text}-to-{Image} {Generation} with
  {Improved} {CLIP}+{GAN} {Space} {Optimization}.
\newblock \emph{arXiv:2112.01573 [cs]}, December 2021.
\newblock URL \url{http://arxiv.org/abs/2112.01573}.
\newblock arXiv: 2112.01573.

\bibitem[Lu et~al.(2019)Lu, Batra, Parikh, and Lee]{lu2019vilbert}
Lu, J., Batra, D., Parikh, D., and Lee, S.
\newblock Vilbert: Pretraining task-agnostic visiolinguistic representations
  for vision-and-language tasks.
\newblock \emph{arXiv preprint arXiv:1908.02265}, 2019.

\bibitem[Lu et~al.(2018)Lu, Wu, Tai, and Tang]{luImageGenerationSketch2018}
Lu, Y., Wu, S., Tai, Y.-W., and Tang, C.-K.
\newblock Image {Generation} from {Sketch} {Constraint} {Using} {Contextual}
  {GAN}.
\newblock In Ferrari, V., Hebert, M., Sminchisescu, C., and Weiss, Y. (eds.),
  \emph{Computer {Vision} – {ECCV} 2018}, volume 11220, pp.\  213--228.
  Springer International Publishing, Cham, 2018.
\newblock ISBN 978-3-030-01269-4 978-3-030-01270-0.
\newblock \doi{10.1007/978-3-030-01270-0_13}.
\newblock URL \url{http://link.springer.com/10.1007/978-3-030-01270-0_13}.
\newblock Series Title: Lecture Notes in Computer Science.

\bibitem[Mirza \& Osindero(2014)Mirza and Osindero]{mirza2014conditional}
Mirza, M. and Osindero, S.
\newblock Conditional generative adversarial nets.
\newblock \emph{arXiv preprint arXiv:1411.1784}, 2014.

\bibitem[Miyato \& Koyama(2018)Miyato and Koyama]{miyato2018cgans}
Miyato, T. and Koyama, M.
\newblock cgans with projection discriminator.
\newblock \emph{arXiv preprint arXiv:1802.05637}, 2018.

\bibitem[Nichol et~al.(2021)Nichol, Dhariwal, Ramesh, Shyam, Mishkin, McGrew,
  Sutskever, and Chen]{nichol2021glide}
Nichol, A., Dhariwal, P., Ramesh, A., Shyam, P., Mishkin, P., McGrew, B.,
  Sutskever, I., and Chen, M.
\newblock Glide: Towards photorealistic image generation and editing with
  text-guided diffusion models.
\newblock \emph{arXiv preprint arXiv:2112.10741}, 2021.

\bibitem[Radford et~al.(2021)Radford, Kim, Hallacy, Ramesh, Goh, Agarwal,
  Sastry, Askell, Mishkin, Clark, Krueger, and
  Sutskever]{radfordLearningTransferableVisual2021}
Radford, A., Kim, J.~W., Hallacy, C., Ramesh, A., Goh, G., Agarwal, S., Sastry,
  G., Askell, A., Mishkin, P., Clark, J., Krueger, G., and Sutskever, I.
\newblock Learning {Transferable} {Visual} {Models} {From} {Natural} {Language}
  {Supervision}.
\newblock \emph{arXiv:2103.00020 [cs]}, February 2021.
\newblock URL \url{http://arxiv.org/abs/2103.00020}.
\newblock arXiv: 2103.00020.

\bibitem[Sangkloy et~al.(2016)Sangkloy, Burnell, Ham, and
  Hays]{sangkloy2016sketchy}
Sangkloy, P., Burnell, N., Ham, C., and Hays, J.
\newblock The sketchy database: learning to retrieve badly drawn bunnies.
\newblock \emph{ACM Transactions on Graphics (TOG)}, 35\penalty0 (4):\penalty0
  1--12, 2016.

\bibitem[Santurkar et~al.(2019)Santurkar, Tsipras, Tran, Ilyas, Engstrom, and
  Madry]{santurkar2019image}
Santurkar, S., Tsipras, D., Tran, B., Ilyas, A., Engstrom, L., and Madry, A.
\newblock Image synthesis with a single (robust) classifier.
\newblock \emph{arXiv preprint arXiv:1906.09453}, 2019.

\bibitem[Stanley(2007)]{stanleyCompositionalPatternProducing2007}
Stanley, K.~O.
\newblock Compositional pattern producing networks: {A} novel abstraction of
  development.
\newblock \emph{Genetic Programming and Evolvable Machines}, 8\penalty0
  (2):\penalty0 131--162, June 2007.
\newblock ISSN 1389-2576, 1573-7632.
\newblock \doi{10.1007/s10710-007-9028-8}.
\newblock URL \url{http://link.springer.com/10.1007/s10710-007-9028-8}.

\bibitem[Sylvain et~al.(2019)Sylvain, Petrini, and Hjelm]{sylvain2019locality}
Sylvain, T., Petrini, L., and Hjelm, D.
\newblock Locality and compositionality in zero-shot learning.
\newblock \emph{arXiv preprint arXiv:1912.12179}, 2019.

\bibitem[Vaswani et~al.(2017)Vaswani, Shazeer, Parmar, Uszkoreit, Jones, Gomez,
  Kaiser, and Polosukhin]{vaswaniAttentionAllYou2017}
Vaswani, A., Shazeer, N., Parmar, N., Uszkoreit, J., Jones, L., Gomez, A.~N.,
  Kaiser, L., and Polosukhin, I.
\newblock Attention {Is} {All} {You} {Need}.
\newblock \emph{arXiv:1706.03762 [cs]}, December 2017.
\newblock URL \url{http://arxiv.org/abs/1706.03762}.
\newblock arXiv: 1706.03762.

\bibitem[Xiang et~al.(2021)Xiang, Liu, Yang, Zhu, Shen, and
  Allebach]{xiangAdversarialOpenDomain2021}
Xiang, X., Liu, D., Yang, X., Zhu, Y., Shen, X., and Allebach, J.~P.
\newblock Adversarial {Open} {Domain} {Adaptation} for {Sketch}-to-{Photo}
  {Synthesis}.
\newblock \emph{arXiv:2104.05703 [cs]}, December 2021.
\newblock URL \url{http://arxiv.org/abs/2104.05703}.
\newblock arXiv: 2104.05703.

\bibitem[Xu et~al.(2018)Xu, Zhang, Huang, Zhang, Gan, Huang, and
  He]{xu2018attngan}
Xu, T., Zhang, P., Huang, Q., Zhang, H., Gan, Z., Huang, X., and He, X.
\newblock Attngan: Fine-grained text to image generation with attentional
  generative adversarial networks.
\newblock In \emph{Proceedings of the IEEE conference on computer vision and
  pattern recognition}, pp.\  1316--1324, 2018.

\bibitem[Zhu et~al.(2017{\natexlab{a}})Zhu, Park, Isola, and
  Efros]{zhu2017unpaired}
Zhu, J.-Y., Park, T., Isola, P., and Efros, A.~A.
\newblock Unpaired image-to-image translation using cycle-consistent
  adversarial networks.
\newblock In \emph{Proceedings of the IEEE international conference on computer
  vision}, pp.\  2223--2232, 2017{\natexlab{a}}.

\bibitem[Zhu et~al.(2017{\natexlab{b}})Zhu, Park, Isola, and
  Efros]{zhuUnpairedImagetoImageTranslation2017}
Zhu, J.-Y., Park, T., Isola, P., and Efros, A.~A.
\newblock Unpaired {Image}-to-{Image} {Translation} {Using}
  {Cycle}-{Consistent} {Adversarial} {Networks}.
\newblock In \emph{2017 {IEEE} {International} {Conference} on {Computer}
  {Vision} ({ICCV})}, pp.\  2242--2251, Venice, October 2017{\natexlab{b}}.
  IEEE.
\newblock ISBN 978-1-5386-1032-9.
\newblock \doi{10.1109/ICCV.2017.244}.
\newblock URL \url{http://ieeexplore.ieee.org/document/8237506/}.

\end{thebibliography}
\bibliographystyle{icml2022}



\end{document}